\pdfoutput=1

\documentclass[11pt,breaklinks,backref=page,colorlinks=true]{article}

\usepackage[final]{acl}

\usepackage{times}
\usepackage{latexsym}

\usepackage[T1]{fontenc}

\usepackage[utf8]{inputenc}

\usepackage{microtype}

\usepackage{inconsolata}

\usepackage{graphicx}
\usepackage{algorithm}
\usepackage[noend]{algpseudocode}
\usepackage{markdown}
\usepackage{spverbatim}

\usepackage{url}
\usepackage{graphicx}
\usepackage{subcaption}
\usepackage{caption}
\usepackage{booktabs}

\renewcommand*{\backref}[1]{}
\renewcommand*{\backrefalt}[4]{%
  \ifcase #1
    Not cited.%
  \or
    Cited on page #2.%
  \else
    Cited on pages #2.%
  \fi
}

\newcommand{\astro}[0]{Astro}
\newcommand{\values}[0]{Values}
\newcommand{\media}[0]{Frame}
\newcommand{\mini}[0]{\texttt{gpt-4o-mini}}
\newcommand{\llama}[0]{\texttt{llama-3.1-8B}}

\usepackage[noabbrev]{cleveref}
\usepackage[group-separator={,}]{siunitx}
\crefname{section}{\S}{\S\S}
\Crefname{section}{\S}{\S\S}
\crefname{table}{Tab.}{}
\crefname{figure}{Fig.}{}
\crefname{algorithm}{Algorithm}{}
\crefname{equation}{eq.}{}
\crefname{appendix}{App.}{}
\crefname{thm}{Theorem}{}
\crefname{prop}{Proposition}{}
\crefname{cor}{Corollary}{}
\crefname{observation}{Observation}{}
\crefname{assumption}{Assumption}{}
\crefformat{section}{\S#2#1#3}
\usepackage[]{hyperref}
\usepackage[noabbrev]{cleveref}

\newcommand{\Sref}[1]{\S\ref{#1}}
\newcommand{\hicode}[0]{\texttt{HICode}}

\title{HICode: Hierarchical Inductive Coding with LLMs}

\author{Mian Zhong\thanks{Equal Contribution} \and Pristina Wang\footnotemark[1]  \and Anjalie Field \\
        Johns Hopkins University \\ \texttt{mzhong8@jh.edu}}

\begin{document}
\maketitle
\begin{abstract}
Despite numerous applications for fine-grained corpus analysis, researchers continue to rely on manual labeling, which does not scale, or statistical tools like topic modeling, which are difficult to control.
We propose that LLMs have the potential to scale the nuanced analyses that researchers typically conduct manually to large text corpora. To this effect, inspired by qualitative research methods, we develop HICode\footnote{Code is available at \url{https://github.com/mianzg/HICode}}, a two-part pipeline that first inductively generates labels directly from analysis data and then hierarchically clusters them to surface emergent themes.
We validate this approach across three diverse datasets by measuring alignment with human-constructed themes and demonstrating its robustness through automated and human evaluations.
Finally, we conduct a case study of litigation documents related to the ongoing opioid crisis in the U.S., revealing aggressive marketing strategies employed by pharmaceutical companies and demonstrating HICode's potential for facilitating nuanced analyses in large-scale data.

\end{abstract}

\section{Introduction}
There are numerous applications for targeted analysis of large text corpora, such as conducting nuanced literature reviews~\citep{blodgett-etal-2020-language,field-etal-2021-survey,Yates2023familybusiness, bowman2023healthcarehci,johnston2025ems}, deeply analyzing trends in social media data~\citep{Lauber2021bigfood} or investigating industry archives like regulatory filings~\citep{Eijk2023tobacco,enache2025lease, wood2024regulatory}.
Researchers and practitioners typically use one of two methods. In the first, they downsample the data to a small enough subset to review manually and conduct \textit{thematic analysis} or \textit{inductive coding}, in which they manually read through the data, label relevant content, and iteratively group labels and re-code the data. This type of data analysis is frequently used by qualitative researchers, mostly commonly for analyzing interview data \citep{Hsieh2005,Thomas2006}, and while it facilitates deep analysis, it does not scale to larger datasets. For example, \citet{Birhane2022mlvalues} use this approach to analyze values encoded in machine learning research, but scalability forces them to limit their analysis 100 highly-cited papers.

Alternatively, researchers use combinations of exploratory text analysis methods, such as lexicon scores, off-the-shelf sentiment models, and most commonly topic models. As an example, \citet{antoniak2019narrative} take this approach in analyzing online birth stories. Despite massive shifts in NLP model capabilities, including increasing performance on complex tasks like math reasoning~\citep{yang2024qwen25math} and code generation~\citep{peng2023copilot}, innovation in corpus analysis has been limited. Latent Dirichlet Allocation (LDA)~\citep{Blei_Ng_Jordan_2003} topic models remain a go-to approach, with evidence that they outperform neural alternatives \citep{hoyle2021automated,hoyle-etal-2022-neural}.
What innovation has occurred has retained existing paradigms like topic modeling and sought to show incremental improvements in metrics like topic coherence \citep{pham-etal-2024-topicgpt}. As a consequence, these approaches also retain the fundamental limitations of these paradigms, including lack of controllability, which makes them unsuited to targeting particular research questions or analysis dimensions.

Rather than retaining topic modeling paradigms, this work revisits the original goals behind corpus analysis: discovering patterns from large corpora. We draw inspiration from qualitative methodology in the humanities and social sciences to shift the approach.
More specifically, we propose that LLMs offer an avenue for scaling the targeted nuanced qualitative research methods to larger data sets, i.e., conducting inductive coding at scale.

To accomplish this, we develop \hicode{}: an LLM-based pipeline with two primary modules, one that generates data labels and one that hierarchically clusters labels. We validate this approach by evaluating its ability to recover human-constructed data labels across three diverse data sets. We further conduct performance and ablation studies, contrasting our hierarchical approach to a more human-like incremental one, and demonstrating that results remain consistent across a range of LLMs.

Finally, we demonstrate the usefulness of our method in a case study analyzing an archive of litigation documents~(3K documents parsed into 160K segments) related to the ongoing opioid epidemic, uncovering aggressive marketing strategies used by pharmaceutical companies~\citep{Caleb2022oida, Eisenkraft2024salesforce}. Overall, by facilitating analyses that are both nuanced and involve large-scale data, \hicode{}---or follow-up methods targeting the same goals---has the potential to enable a broad array of previously infeasible studies, advancing computational social science, digital humanities, and other fields involving analyses of large-scale text corpora.

\section{Background: What is inductive coding?}
In inductive coding, data labels are derived directly from data. These labels are further grouped or analyzed as meaningful themes to answer an exploratory research question like, what tactics and incentives have opioid manufacturers used to increase sales? \citep{Eisenkraft2024salesforce}. The process is commonly applied in qualitative data analysis, such as to derive findings from interview studies \citep{Thomas2006}. \textit{Inductive} data coding contrasts \textit{deductive} coding, in which data labels are drawn from a pre-existing theory or codebook rather than derived directly from the data. Deductive coding is more common in NLP and has been previously investigated as an application for LLMs \citep{Xiao2023,ziems-etal-2024-large}.
Inductive coding also contrasts with topic modeling. While both methods seek to derive patterns directly from raw data, topic modeling is typically unsupervised, rather than targeted towards a particular research question or analysis dimensions. For example, the topic of a paper is very different from its encoded values \citep{Birhane2022mlvalues}.
Lastly, inductive coding differs from summarization in its focus on particular parts of documents that may be unimportant for general summaries and its construction of themes across documents.

\begin{figure*}[hbt!]
    \centering
   \includegraphics[width=\linewidth]{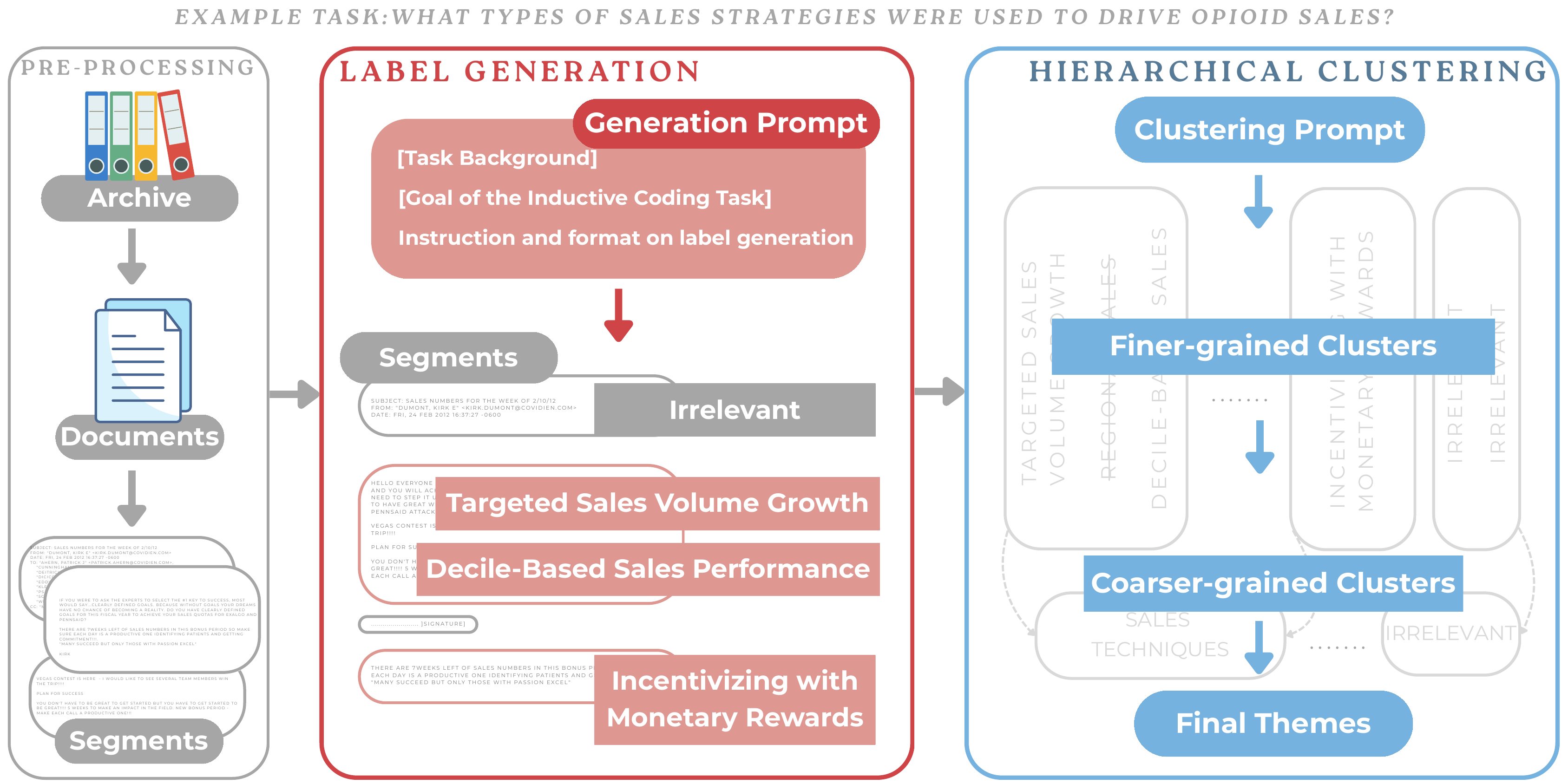}
    \caption{\hicode\ pipeline overview. Given text segments, the pipeline runs (1) Label Generation module to identify relevant segments and generate label(s) corresponding to a provided goal of inductive coding, and (2) Hierarchical Clustering module to derive insightful themes from the generated labels.}
    \label{fig:enter-label}
\end{figure*}

\section{Methodology}
We intentionally construct a pipeline, \hicode{}, that is suited to scalable automated analysis rather than mimicking the incremental and iterative way that humans typically annotate data, which we discuss for comparison in \Sref{sec:comparison_models}.
\hicode{} consists of two primary modules that run sequentially: one that generates labels for each data point and one that merges and clusters the labels across data points. Prior to the modules, if the input is long-text like a litigation document, the documents are parsed into text segments~(e.g., paragraph-level) for better fine-grained label generation. We first introduce these modules and then discuss the motivation behind this approach.

\paragraph{Label Generation}
The goal of the label generation module is to produce clear and concise labels for each text input that are relevant to a research question or analysis dimension that a user cares. To accomplish this, we craft an LLM prompt with two user-provided components: (1) a short description of background information, for example, in analyzing \textit{what are the encoded values in machine learning research?}, the definition of ``encoded values'' needs to be specified; (2) the goal of the inductive coding task. These parts of the prompt are expected to change across datasets, and are necessary to allow a researcher to direct the model to focus on a particular research question, as opposed to producing generic topics.
Then, regardless of the dataset, the next part of the prompt directs the model to identify relevance and generate label(s) that are observational, concise, and clear. Multiple labels are allowed for a given segment.
The outputted labels are the fine-grained initial codes towards conducting the analysis. 

\paragraph{Hierarchical Clustering}
Once initial labels are generated, the clustering module hierarchically groups them and distills abstract, insightful and meaningful themes.
We implement this clustering through repeated rounds of LLM prompting, which our initial experiments found to be more reliable than traditional clustering methods.

Specifically, we randomly divide the generated labels into batches of size $100$. We then prompt an LLM with the goal of the inductive coding task defined in the generation module, a batch of generated labels, and a fixed instruction to synthesize similar labels into themes.
Once all initial batches have been processed, the outputted themes become inputs for the next iteration of clustering. We repeat this process until reaching a pre-determined maximum number of iterations or convergence to a threshold of number of themes, as specified by the user.

\paragraph{Pipeline Motivation}

\hicode{} has the benefit that label generation for each text segment and each batch of clustering is entirely independent, which allows running modules in parallel to efficiently process large datasets. Additionally, because these two modules are entirely distinct, this pipeline facilitates future follow-up work on distilling smaller specialized models for each module, rather than requiring all-purpose LLMs.
Finally, producing fine-grained codes that are later clustered gives more control to the user: a user may start with initial high-level themes and then walk back to earlier iterations of clustering in order to identify detailed labels for themes of interest. We demonstrated this process in our case study (\Sref{sec:case_study}).

\section{Evaluation Metrics}
As our primary task involves inductively constructing a labeling scheme, rather than assigning labels under a known scheme, there are no clear pre-existing evaluation metrics. Instead we design novel automated metrics to compare themes produced by \hicode{} with human-annotated data.
As qualitative data analyses can vary widely and are accepted to involve human interpretation \citep{braun2006using,ridder2014qualitative}, we design metrics that measure closeness to human labels with a range of tolerance to facilitate model comparisons, rather than expecting exact matching.

An ideal set of predicted inductive themes would be both \textit{comprehensive} (include all the same themes that people found) and \textit{minimal} (not contain many extra themes). Furthermore, generated themes should be true to the underlying text segments: given that a predicted theme matches a human one, the data labeled with that theme should be the same.
To this end, we calculate theme-level and segment-level metrics which are modified precision and recall scores.

\paragraph{Notation}
We denote the set of gold themes annotated by humans as $G$, the set of final clustering themes as $T$, and the set of all text segments to be labeled as $S$.

\subsection{Theme-level Precision and Recall}
We use off-the-shelf embedding models to compute embedding representations for each $g \in G$ and $t \in T$. We then compute cosine similarity of embeddings for every pair $(g,t)$. If their similarity is above a given threshold $k$, we consider the pair matched. There can be multiple model themes matched to one gold theme and vice versa, which allows for differing levels of granularity in annotations. 
Denoting the gold themes that get matched as $G_{match}\subseteq G$ and similarly $T_{match}\subseteq T$, we define the theme-level precision as $\frac{|T_{match}|}{|T|}$ and recall as $\frac{|G_{match}|}{|G|}$.
We report all metrics at different values of $k$, thus allowing flexibility in determining how similar the generated themes and gold themes are expected to be.

\subsection{Segment-level Precision and Recall}
For a set of matched themes, segment-level metrics are a weighted average of the per-theme metrics.
\paragraph{Precision} For each model theme $t$ in $T_{match}$, let the set of segments labeled as this theme be $S_{t}$. 
The precision is the proportion of these segments that are also labeled as equivalent gold theme, denoted as $S_t^{matched}$, i.e., $Prec_t = \frac{|S_t^{matched}|}{|S_t|}$.
The overall segment-level precision is the summation of the weighted precisions. For
$w_{t}=\frac{|S_t|}{\sum_t|S_t|}$,

\[Precision_{seg}=\sum w_t\cdot Prec_t\]

\paragraph{Recall}
Similarly, for each gold theme $g$ in $G_{match}$, the set of the overlapping text segments that are both labeled as $g$ and its equivalent model theme is denoted as $S_g^{matched} \subseteq S_g$, where $S_g$ is all segments labeled as $g$. We have the overall segment-level recall as follows,
\[Recall_{seg}=\sum w_g\cdot Recall_g,\]
where  $Recall_g = \frac{|S_g^{matched}|}{|S_g|}$ and 
$w_{g}=\frac{|S_g|}{\sum_g|S_g|}$.

\section{Experiments}

\subsection{Comparison Models}
\label{sec:comparison_models}

We compare our proposed hierarchical approach with an alternative incremental approach and two existing baselines, TopicGPT~\citep{pham-etal-2024-topicgpt} and LLooM~\citep{lam-etal-2024-lloom}.

\paragraph{Incremental Label Generation}
In conventional content analysis, researchers often take an incremental approach to data coding. They draft a preliminary set of labels based on a sample of the data. Then they annotate the remaining data starting with these initial labels, adding new labels when they find data that does not fit into an existing one, and recoding data as needed \citep{Hsieh2005}. 
Thus, we design an incremental LLM-pipeline that mimics this process for comparison, which conducts generation, merging, and dropping of labels to develop final themes.
The distinction between hierarchically merging data labels after processing all of the data and incrementally updating the labeling scheme during data processing has also been explored in book-length summarization \citep{chang2024booookscore}.
One iteration of the incremental pipeline first generates labels for a random sample of unseen data, second, merges these labels into higher-level themes, and third, drops themes that have been assigned to a low number of segments for several rounds.
The full incremental pipeline is repeated for several iterations where later iterations skip generating new labels, just conducting merging and dropping.
The incremental approach fully stops when all labels have numbers of segments that are above the threshold.

\paragraph{TopicGPT} TopicGPT \citep{pham-etal-2024-topicgpt} is an LLM-prompting based pipeline for topic modeling. It uses a uniformly sampled subset of the dataset to generate topics, where the generation is done incrementally by prompting the model to assign an existing label or produce a new one for each text input. The pipeline requires providing an initial seed topic and its description.
In our experiments, we use a real example theme for each dataset as the seed.
TopicGPT then assigns the generated topics to the entire dataset with an assignment labeling step, also accomplished through LLM-prompting.

\paragraph{LLooM} LLooM \citep{lam-etal-2024-lloom} is an LLM-prompting framework designed for ``concept induction''. 
LLooM conducts multiple data passes to distill, cluster, synthesize and assign generated concepts in an interactive workbench. While the end-goals of LLooM are more similar to our end-goals than TopicGPT, the system was not evaluated on inductively coded data, and its design better facilitates smaller-scale interactive exploration than targeted corpus analysis.

\paragraph{Pipeline Implementation}

For fair comparison, we use \texttt{gpt-4o-mini} in all modules of all pipelines, except where otherwise specified for ablation experiments~(\cref{sec: ablation}) for which both proprietary and open-sourced LLMs with different model sizes are compared.
The example topics for TopicGPT, seed words for LLooM, and the prompts \hicode{} and incremental approaches can be found in \cref{app: prompts} and \cref{app:experiments}.

\subsection{Datasets}

We evaluate models for their ability to re-create three human-labeled datasets. We select these datasets to cover a range of disciplines, types of text data, and the size of the inductively coded label set as summarized in~\cref{tab:data-summary}. 
\begin{table}[htb!]
\centering
\resizebox{\linewidth}{!}{
\begin{tabular}{crrrrr}
\toprule
data & \# themes & \# doc & \# seg & \begin{tabular}[c]{@{}c@{}}multi-\\labeled?\end{tabular} & \begin{tabular}[c]{@{}c@{}}avg seg\\length\end{tabular}\\
\midrule
Frame & 15 & 11903 & 112585 & Yes & 164.21\\
Astro  & 9 & 369 & 369 & No & 100.95\\
Values & 82 & 100 & 2157 & Yes & 172.83\\
OIDA & N/A & 3861 & 163173 & N/A & 306.51\\

 \bottomrule
\end{tabular}}
\caption{
Summary statistics for evaluation data sets and the unannotated OIDA data used in our case study. Unit of average length is \# characters.
}
\label{tab:data-summary}
\end{table}
\paragraph{Media Frames Corpus~(Frame)} The Media Frames Corpus is a dataset of news articles annotated for policy frames, such as ``Economic'' or ``Morality''~\citep{Card2015mediaframe}. There are two primary limitations to conducting evaluations with this dataset. First, coding scheme was not derived entirely inductively  \citep{boydstun2014tracking}, and second, as the dataset has existed since 2015, there is a risk that LLMs may have been exposed to it in pre-training data.
Despite these limitations, we choose this corpus because the annotation scheme was designed to cross-cut policy issues, meaning the framing labels are intentionally distinct from topics. This corpus has also been widely used in NLP literature, whereas our other data sets have not previously been used for evaluations.
We parse each document into paragraphs as the input segments for our experiments.

\paragraph{Astro Queries~(Astro)} We use a dataset of queries sent to an LLM-powered bot designed to aid astronomers in interacting with astronomy literature~\citep{hyk2025from}.
This data was inductively coded to determine the types of queries users sent to the model, such as ``Knowledge seeking: Specific factual'' or ``Stress Testing'', where the goal was to analyze evaluation strategies that users employed in testing the bot. Unlike other datasets, this data was \textit{entirely} inductively coded. Furthermore, the data was not publicly released by the knowledge cut-off date of any of the models we evaluate.
We directly use the original queries as the input segments.

\paragraph{ML Values~(Values)} We use the dataset of 100 machine learning research papers annotated for encoded values from~\citet{Birhane2022mlvalues}. The data consists of manually selected snippets from articles annotated with values like ``Efficiency'', ``Performance'', and ``Privacy'' using a hybrid deductive and inductive coding approach. While values originally determined deductively may be difficult for our models to capture, given the fine-grained annotations and semi-inductive approach, we expect many of them to be recoverable. This dataset contains the largest number of themes of any of the datasets we use, thus facilitating more robust evaluation. We directly use selected snippets as input segments.
 
\begin{table*}[htb!]
\centering

\resizebox{0.9\linewidth}{!}{
\begin{tabular}{clcccccc}
\toprule
  & & \multicolumn{2}{c}{\textbf{k=0.4}} & \multicolumn{2}{c}{\textbf{k=0.45}} & \multicolumn{2}{c}{\textbf{k=0.5}} \\
  \midrule
  data & pipeline & Precision & Recall  & Precision & Recall & Precision & Recall \\
  \midrule
   Values & TopicGPT & 0.88 $(\pm 0.07)$ & 0.52 $(\pm 0.12)$ & 0.71 $(\pm 0.15)$ & 0.31 $(\pm 0.07)$ & 0.58 $(\pm 0.15)$ & 0.24 $(\pm 0.06)$ \\
  & LLooM & 0.62 $(\pm 0.21)$ & 0.34 $(\pm 0.13)$ & 0.54 $(\pm 0.23)$ & 0.21 $(\pm 0.09)$ & 0.33 $(\pm 0.15)$ & 0.15 $(\pm 0.07)$ \\ 
 & Incremental & 0.92 $(\pm 0.10)$ & 0.27 $(\pm 0.12)$ & 0.71 $(\pm 0.18)$ & 0.18 $(\pm 0.07)$ & 0.56 $(\pm 0.21)$ & 0.11 $(\pm 0.05)$ \\
 
 & \hicode{} & \textbf{0.96} $(\pm 0.05)$ & \textbf{0.57} $(\pm 0.04)$ & \textbf{0.83} $(\pm 0.07)$ & \textbf{0.40} $(\pm 0.04)$ & \textbf{0.62} $(\pm 0.12)$ & \textbf{0.27} $(\pm 0.04)$ \\
 \midrule
 
 Astro & TopicGPT & 0.04 $(\pm 0.02)$ & 0.49 $(\pm 0.12)$ & 0.04 $(\pm 0.02)$ & 0.47 $(\pm 0.06)$ & 0.03 $(\pm 0.01)$ & 0.33 $(\pm 0.00)$ \\
  & LLooM & 0.17 $(\pm 0.26)$ & 0.24 $(\pm 0.31)$ & 0.07 $(\pm 0.14)$ & 0.13 $(\pm 0.23)$ & 0.03 $(\pm 0.07)$ & 0.07 $(\pm 0.19)$ \\
 & Incremental & 0.19 $(\pm 0.05)$ & \textbf{0.67} $(\pm 0.10)$ & 0.10 $(\pm 0.03)$ & \textbf{0.53} $(\pm 0.15)$ & 0.07 $(\pm 0.04)$ & \textbf{0.40} $(\pm 0.21)$\\
 & \hicode{} & \textbf{0.53} $(\pm 0.17)$ & 0.51 $(\pm 0.08)$ & \textbf{0.22} $(\pm 0.24)$ & 0.33 $(\pm 0.29)$ & \textbf{0.08} $(\pm 0.14)$ & 0.18 $(\pm 0.30)$ \\

 \midrule
 Frame & TopicGPT & \textbf{0.82} $(\pm 0.13)$ & 0.76 $(\pm 0.09)$ & \textbf{0.71} $(\pm 0.06)$ & 0.64 $(\pm 0.15)$ & 0.49 $(\pm 0.08)$ & \textbf{0.51} $(\pm 0.13)$\\
   & LLooM & 0.49 $(\pm 0.19)$ & 0.48 $(\pm 0.33)$ & 0.37 $(\pm 0.10)$ & 0.40 $(\pm 0.30)$ & 0.31 $(\pm 0.12)$ & 0.29 $(\pm 0.21)$\\
 & Incremental & 0.78 $(\pm 0.08)$ & 0.68 $(\pm 0.07)$ & 0.69 $(\pm 0.10)$ & 0.56 $(\pm 0.05)$ & \textbf{0.50} $(\pm 0.16)$ & 0.43 $(\pm 0.14)$\\
 & \hicode{} & 0.68 $(\pm 0.10)$ & \textbf{0.81} $(\pm 0.22)$ & 0.54 $(\pm 0.13)$ & \textbf{0.67} $(\pm 0.25)$ & 0.41 $(\pm 0.10)$ & 0.49 $(\pm 0.19)$\\ 

 \bottomrule\\
\end{tabular}}
\caption{
Theme-level Precision and recall scores on TopicGPT (without removing the seed topic), LLooM, Incremental and \hicode{} using \texttt{gpt-4o-mini}. The results are the average out of 5 runs for each pipeline, and the similarity threshold $k$ is set to $0.4, 0.45, 0.5$ to select matched pairs of the gold and model output themes. Parentheses $()$ indicate the confidence interval range by using the t-distribution.
}
\label{tab:model-comparison}
\end{table*}

\section{Results}
\subsection{Overall Performance}\label{subsec:overall-performance}
We compare the performance of \hicode{} with TopicGPT, LLooM, and the incremental approach in \cref{tab:model-comparison} on theme-level precision and recall scores.\footnote{TopicGPT metrics are slightly inflated because the model output contains a gold theme from setting the seed to guide topic generation}

Over the \values\ dataset, which has the largest number of themes, \hicode{} achieves the best precision and recall. The precision under the most lenient matching threshold ($k=0.4$) is quite high ($0.96$), indicating that nearly all themes identified by the method were similar to the human-labeled themes, though the lower precision at stricter thresholds indicates they did not always match exactly.
Over \astro, the incremental and \hicode{} methods both far out-perform TopicGPT and LLooM. While the incremental method has higher recall ($0.67 @ k=0.4$), the hierarchical method has comparable recall ($0.51$) with much better precision ($0.53$ vs. $0.19$).
TopicGPT does perform well on \media\ likely because framing annotations in this dataset are \textit{emphasis} frames \citep{chong2007framing} that are very topic-like and a gold frame is provided to TopicGPT. Nevertheless, the \hicode{} pipeline maintains comparable or better recall.
In contrast, on \astro\, which has a harder goal of inductive coding to focus on the ``query types'' rather than ``query content'', TopicGPT's performance degrades much and our hierarchical and incremental pipelines have the best results.
Furthermore, as the clustering is done hierarchically, we have the flexibility to control the granularity of themes. Therefore, we likely could improve recall by choosing finer-grained clustering results.

\subsection{Human evaluation of \astro\ }
As seen in \cref{tab:model-comparison}, while the automated metrics are useful for facilitating comparisons of models, the exact values vary depending on the choice of $k$, and thus are limited in their absolute reflection of performance. Hence, we more qualitatively evaluate predictions and conduct a human evaluation study over \astro\ for \hicode{} and TopicGPT. We focus on this dataset because \cref{tab:model-comparison} suggests it was the most difficult with the lowest automated metrics for all models. It is also the only dataset that was entirely inductively coded with least risk of leakage into LLM pre-training data.

\begin{figure*}[htb!]
    \centering
         \includegraphics[width=\textwidth]{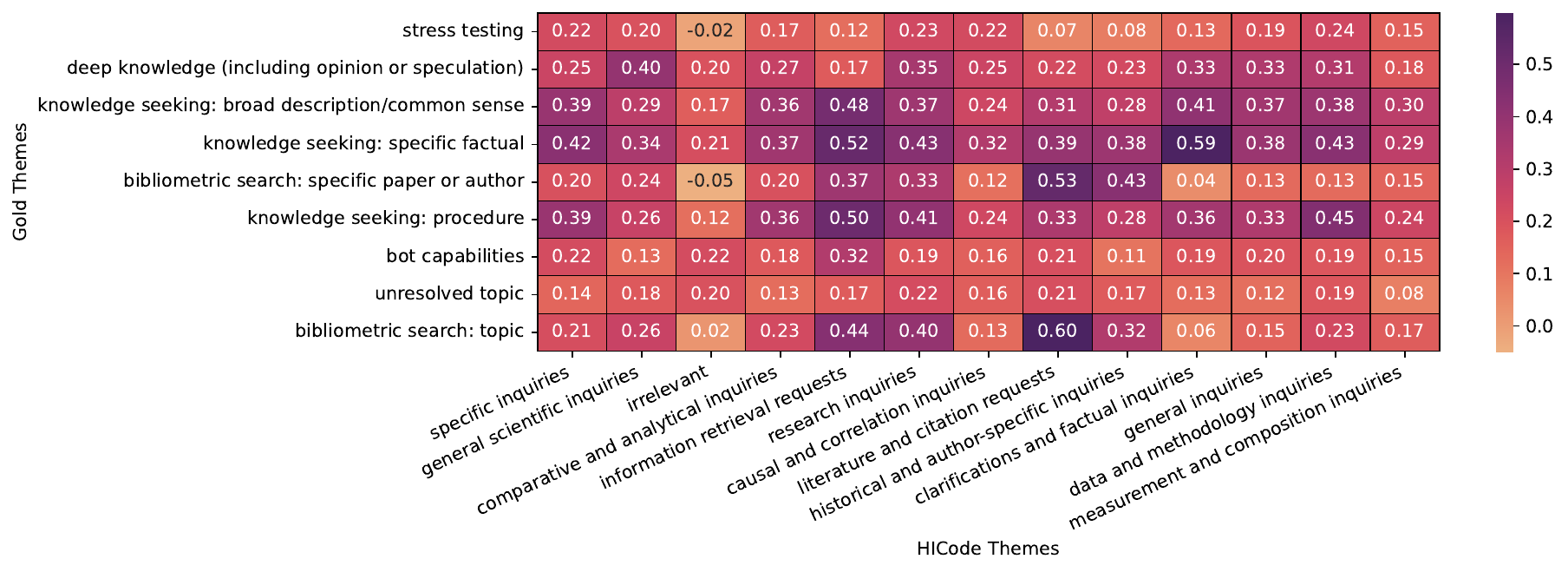}
        \caption{Heatmap comparing themes identified by the \hicode{} (horizontal) with the gold human-labeled themes (vertical) for \astro. Each box contains the automatically estimated similarity score between the predicted and gold theme, with darker colors signifying higher similarity. The generated themes overall do target similar concepts as the original themes, but granularity differs.}
        \label{fig:automated-hierarchical}
    \label{fig:heatmap}
\end{figure*}

In \cref{fig:heatmap} we show a heatmap that uses the automatically inferred similarity scores to visualize how themes generated by \hicode{} compare to gold themes. The predicted themes generally do reflect the type of query posed by the user (e.g., ``general scientific inquiries''), thus matching the target analysis dimension, in contrast to themes predicted by TopicGPT reported in \cref{sec:heatmap}, which focus on the content of the query (e.g., ``exoplanet research''). 
The generated themes do have different granularity levels than the gold themes. For example, the generated ``information retrieval requests'' partially maps to all of the ``knowledge seeking'' gold themes, and similarly, the gold scheme subdivides ``bibliometric search'' into two themes, which both map to one generated theme: ``literature and citation requests''. 
\cref{fig:heatmap} also suggests that automated metrics may under-count matches. For example, ``general scientific inquiries'' partially recovers the gold theme ``knowledge seeking: broad description'' but the similarity score is only $0.29$, which would count as unmatched for all the $k$ values we report in \cref{tab:model-comparison}.

As automated metrics are limited in their representation of results, we further conduct a human evaluation by recruiting two of the original annotators of \astro\ to manually compare the generated themes with the human themes. 
We provide outputs from three runs each of TopicGPT and \hicode{} and ask annotators to score each pair of gold and generated themes with $0.5$ if the gold theme is partially recovered by the model~(e.g., gold theme is a superset of the model theme or vice versa) and $1$ if the gold theme exactly matches the generated theme. The resulting agreement rate calculated using Krippendorff’s alpha is $0.31$. We use the annotations by averaging them and define a matched theme if the averaged score $\geq 0.5$.

As shown in \cref{tab: human-eval-astro}, using human judgments, \hicode{} achieves both high precision at $0.72$ and recall at $0.74$, vastly outperforming TopicGPT, which has an even worse recall than under the automated evaluation~($0.49 \rightarrow 0.33$).

\begin{table}[htb!]
\centering

\resizebox{0.85\linewidth}{!}{
\begin{tabular}{lrrrr}
\toprule
  & \multicolumn{2}{c}{\textbf{TopicGPT}}  & \multicolumn{2}{c}{\textbf{\hicode{}}} \\
  \midrule
   Similarity & Prec & Recall  & Prec & Recall \\
  \midrule
 Cosine~($k=0.4$) & 0.04 & 0.49 & 0.53 & 0.51\\

 Human & 0.18 & 0.33 & 0.72 & 0.74 \\
 \bottomrule
\end{tabular}}
\caption{
Theme-level precision and recall for \astro. ``Cosine'' row is the automated metric from \cref{tab:model-comparison} and ``Human'' row is where matching of pipeline output themes to gold themes was manually conducted by two original annotators of the dataset.}
\label{tab: human-eval-astro}
\end{table}

This human evaluation offers strong evidence that \hicode{} gives themes that capture similar information as human annotations, and the metrics in \cref{tab:model-comparison} are a conservative estimate of performance.

\subsection{Segment-level metrics}

We report segment-level metrics for 
\astro\ and \values\ in \cref{tab:model-comparison-segment-3}. We do not include \media\ since this data was not originally annotated on paragraph-level segments but through free selection of span by the annotators.
TopicGPT, LLooM and the incremental approach all involve a second pass over the data to assign the final generated themes back to the text segments. While this re-assignment pass is not strictly necessary for \hicode{}, we conduct it anyway for fair comparison.

\begin{table}[htb!]
\centering

\resizebox{0.9\linewidth}{!}{
\begin{tabular}{clrrrrrr}
\toprule
  & & \multicolumn{2}{c}{\textbf{k=0.4}}  & \multicolumn{2}{c}{\textbf{k=0.5}} \\
  \midrule
  data & pipeline & Prec & Recall  & Prec & Recall\\
  \midrule
 
 Astro & TopicGPT &  0.47 & 0.14 & 0.43 & 0.19  \\
 
  & LLooM &  \underline{0.55} & \underline{0.96}  & \underline{0.50} & \underline{0.90} \\
 
 & Incremental &  0.50 & 0.14  & 0.56 & 0.07 \\
 
 & \hicode{} &  \textbf{0.57} & 0.76 & \underline{\textbf{0.57}} & \underline{\textbf{0.30}} \\

 \midrule
 Values & TopicGPT & \textbf{0.38} & 0.46 & \textbf{0.34} & 0.56\\
 
  & LLooM & 0.18 & 0.15 & 0.13 & 0.16 \\ 
 
 & Incremental & 0.24 & 0.32 & 0.28 & 0.36 \\
 & \hicode{} & 0.34 & \textbf{0.61} & 0.25 & \textbf{0.64} & \\
 \bottomrule\\
\end{tabular}}
\caption{
Segment-level precision and recall scores averaged over 5 runs of each approach. \underline{underlined metrics} indicate some runs produced no matched themes and could not be included: LLooM from 3 out of 5 runs~($k=0.4$) and 1 out of 5 runs~($k=0.5$); \hicode{} from 2 out of 5 runs~($k=0.5$).
}
\label{tab:model-comparison-segment-3}
\end{table}

As segment-level metrics are only computed for sets of matched themes, which vary across each model, metrics are not directly comparable but do provide some visibility into whether each pipeline labels the data correctly.  
\hicode{} has higher segment recalls than other methods.\footnote{with the exception of one $0.96$ recall from LLooM; however, this row only contains results from one run, as the other 4 runs returned an empty set of matched themes, making segment metrics incomputable}
\hicode{} also has the highest precision on \astro.

\subsection{Model Ablation Study}\label{sec: ablation} 
We further conduct an analysis on the impact of different LLM models used in the generation and clustering modules for \hicode{}.
TopicGPT and LLooM are both entirely based on API-based models which are not suitable for private data and may not be cost-effective with larger data scale.
Moreover, ~\citet{pham-etal-2024-topicgpt} find that open-sourced LLMs generate labels poorly.
We conduct ablations on the \values\ data, which has the largest number of themes, using popular open-sourced~(e.g., Llama and Mixtral) and API-based LLMs. 

We report full results in \Cref{sec:app_abalations}. Surprisingly, when we fix \mini\ as the clustering model, the choice of generation model has little impact on performance, as all models achieve precision $\geq 0.53$ and recall $\geq 0.22$ as shown in \cref{tab:generation-abalation}.
When we compare using \llama\ and \mini\ for label generation with varying clustering models, \mini\ (precision range: $0.54-0.64$; recall: $0.26-0.40$) slightly outperforms \llama\ (precision: $0.47-0.58$; recall: $0.21 - 0.33$), which can be found in \cref{tab:clustering-ablation-llama} and \cref{tab:clustering-ablation}. This result suggests that a strong label generation model may be able to compensate for a weaker clustering model.
Nonetheless, we do not observe major performance variation under different model combinations.

\section{Case study: Identifying Sales Strategies in Opioids Industry Documents Archive}
\label{sec:case_study}
\begin{figure*}
    \centering
    \includegraphics[width=0.9\linewidth]{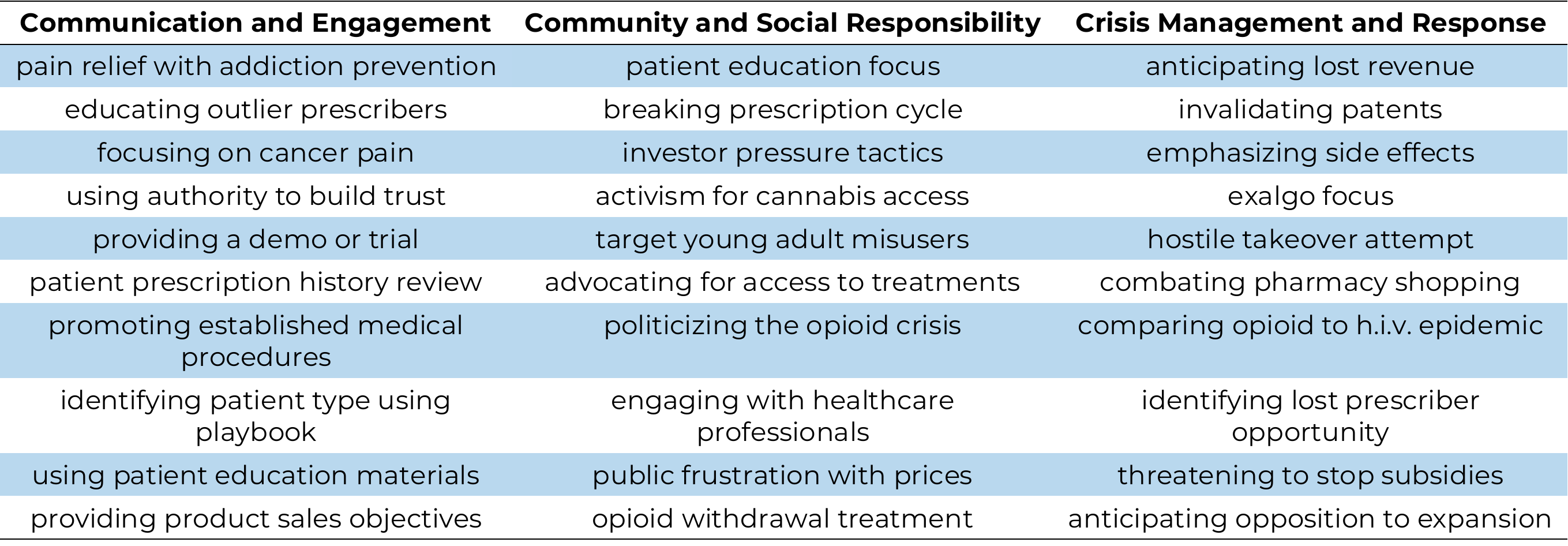}
    \caption{Three themes with randomly selecting cluster labels generated by \hicode{} over OIDA.}
    \label{fig:theme-sample-labels}
\end{figure*}
We conduct a case study to demonstrate how \hicode{} has the potential to facilitate deep analysis of large corpora using the UCSF-JHU Opioid Industry Documents Archive~(OIDA)~\citep{Caleb2022oida}. OIDA is a growing repository containing millions of internal corporate documents related to opioid litigation.
The overdose epidemic in the U.S. has spanned over 20 years and resulted in more than one million deaths, with high involvement of prescription opioids~\citep{Caleb2022oida}. 
Analysis of OIDA has the potential to uncover vital information about ways pharmaceutical companies have contributed to this crisis, such as using marketing strategies that target key opinion leaders~\citep{Gac2024kol} or women and children~\citep{Yakubi_Gac_Apollonio_2022}.
However, the release of data as difficult-to-read unstructured PDFs with OCR converted plain texts~(see \cref{app:case-study}) has restricted these prior investigations to manual investigation of a tiny percentage~(e.g., 200 to 600 documents) of the archive.

We draw inspiration from~\citet{Eisenkraft2024salesforce}, who study \textit{what types of sales strategies or techniques were used to drive Opioid sales?}. They initially searched the archive for ``sales contest'', but found the returned documents too broad and numerous and instead focused on a narrower subset for manual analysis. 
We use \hicode{} to conduct the scaled analysis that was infeasible manually. 
Specifically, we search for ``sales contest'' in  Mallinckrodt email collection, retrieving 3,861 emails OCRs which we parse into 163,173 segments.
The pipeline uses \llama\ in generation and \mini\ for clustering.
In generation stage, \hicode\ labels $40\%$ of the data as irrelevant (65,642 segments with examples in Appendix \cref{fig:irrelevant-quote}).
The clustering ran for five iterations to reach 17 final themes from 70K generated labels~(\cref{fig:salescontest_label_distribution}).
Among the final themes, ``sales strategies and techniques'' dominantly contains the most generated labels~($\approx 14K$) followed by ``regulatory and compliance''~($\approx6K)$. 
As the clustering module is hierarchical, we can further break down the themes and trace back to finer-grained labels. In \cref{fig:theme-sample-labels}, we highlight such fine-grained labels associated with three interesting final themes: ``Communication and Engagement'', ``Crisis Management and Responses'', and ``Community and Social Responsibility''.

First, the labels within ``Communication and Engagement'' suggest a diverse range of sales communication techniques for different customer groups~(e.g., ``identifying patient type using playbook'') or medical conditions~(e.g., ``focusing on cancer pain''). These results further support relevant public health studies. For instance, while \citet{Eisenkraft2024salesforce} identify hypertargeting high-decile prescribers, our analysis additionally identifies efforts for ``educating outlier prescribers'' which labels the following segment: 
``With all the issues with opioid prescribing and potential abuse, you have to understand what your own organization is doing [...] You should benchmark practices and then try to educate prescribers who may be outliers.''\footnote{\url{https://www.industrydocuments.ucsf.edu/opioids/docs/\#id=zskd0237}}

Aside from micro-level engagement, higher-level crisis awareness and prevention is also integral to sales strategies. Investigating documents labeled with the theme ``Crisis Management and Response'' reveals strategizing around ``anticipating lost revenue'' from industry competitors, ``identifying lost prescriber opportunity'' for regional sales monitoring, and ``anticipating opposition to expansion'' due to potential electoral consequences with details in \cref{app:case-study}.

Meanwhile, the theme ``Community and Social Responsibility'' also suggests possible evidence that companies were aware of the public harm of their product.  In reviewing texts labeled under labels like ``breaking prescription cycle'' and ``public frustration with prices'', we find corporate daily newsletters circulating. These internal newsletters offer evidence that employees were informed about news coverage of growing crisis, including content related to prescription drug safety, government affairs, and pharmaceutical corporations and their products.
In particular, the label of ``targeting young adult misusers'' designates circulated news about the vulnerability of and impact to this population. While the intention behind circulation this news is unclear from the documents we reviewed, at a minimum, they do suggest awareness and monitoring of the crisis.

\section{Related Work}

To date, most LLM-driven data annotating tasks in NLP have focused on deductive data coding, classifying text based on a pre-existing annotation scheme \citep{Xiao2023,ziems-etal-2024-large}.
Work that takes a more inductive approach with the goal of open-ended corpus analysis has focused on topic modeling. Initial topic models that leverage pre-trained language models, such as BERTTopic \citep{grootendorst2022bertopic} and Contextualized Topic Models \citep{bianchi-etal-2021-pre}, extract pre-trained embeddings to cluster or incorporate as features. More recently, TopicGPT \citep{pham-etal-2024-topicgpt} consists of a LLM-prompting pipeline for topic modeling that demonstrates strong performance as compared to previous models, which is why we use this model for comparison.

A separate line of work has focused on developing automated tools to assist researchers in qualitative coding, such as  CollabCoder \citep{Gao2024} and Scholarstic \citep{Hong2022} and other systems leveraging LLM suggestions \citep{dai-etal-2023-llm,parfenova-etal-2025-text, pacheco-etal-2023-interactive}. This work has often been conducted by HCI researchers with extensive interface design to facilitate human involvement in the coding process. Most evaluations are conducted over interview datasets, which are necessarily small as they are limited by the number of interviews a research team can conduct. These systems aim to aid qualitative researchers in analyzing data by hand, rather than scaling analyses to conduct corpus analyses. LLooM \citep{lam-etal-2024-lloom}, which focuses on ``concept induction'' is an intermediary between this line of work and our work. While \citet{lam-etal-2024-lloom} do conduct comparisons against topic models on non-interview datasets, they substantially downsample data for evaluation, and their released code was developed to be interactive, requiring us to make modifications for more scaled analyses.

\section{Conclusions}
We proposed \hicode, an LLM-based pipeline for conducting deep nuanced analysis over large-scale data. Our evaluations and case study suggest that this method has high potential for enabling previously infeasible analyses, with opportunities to assist in-depth exploratory corpus analysis.

\section*{Limitations}

The primary limitation of our work is the difficulty of estimating reliable evaluation metrics in this setting. As qualitative data analyses can vary widely and are accepted to involve human interpretation \citep{braun2006using,ridder2014qualitative}, exact replication of human labels is not expected, which makes evaluations difficult. We take steps to mitigate this limitation, including reporting metrics over 5 separate runs for each model and using a range of metrics with varying levels of tolerance. Nevertheless, future work is needed to further improve the reliability of evaluation metrics.

We also conduct evaluations over a specific set of datasets and models. We specifically chose datasets with wide variability in size and domain and we conduct model ablations studies, but we nevertheless cannot conclusively determine how results will generalize to new settings.
Relatedly, our method also relies on LLM-prompting and requires a user to provide background context on their research question or analysis dimension. While we did not do any prompt optimization and only tried one prompt for each dataset--- suggesting that careful prompt crafting may not be needed for the pipeline to work--- results may vary by the user's choice of prompt.

\section*{Acknowledgments}
We gratefully thank Caleb Alexander, Ioana Ciucă, Jie Gao, Kevin Hawkins, Alina Hyk, Louis Hyman, 
Sandesh Rangreji, Brian Wingenroth,  John Wu, and Ziang Xiao for their helpful feedback on this work. This work was supported in part by a Johns Hopkins Discovery Award.

\bibliography{anthology,custom} 

\appendix

\section{Model Prompts}
\label{app: prompts}
In this section, we provide the prompting templates used in our \hicode{} pipelines, where {} is a variable that changes the content depending on the dataset and inductive coding task.

\subsection{Label Generation}
\begin{spverbatim}
{Background Information}

{Goal of Inductive Coding}
Instruction:
- Label the input only when it is HIGHLY RELEVANT and USEFUL for {Goal of Inductive Coding}.
- Then, define the phrase of the label. The label description should be observational, concise and clear.
- ONLY output the label and DO NOT output any explanation.

Format:
- Define the label using the format \"LABEL: [The phrase of the label]\". 
- If there are multiple labels, each label is a new line. 
- If the input is irrelevant, use \"LABEL: [Irrelevant]\". 
- The label MUST NOT exceed 5 words.

\end{spverbatim}

\subsection{Hierarchical Clustering}

\begin{spverbatim}
Synthesize the entire list of labels by clustering similar labels that are inductively labeled. The clustering is to finalize MEANINGFUL and INSIGHTFUL THEMES for {Goal of inductive coding}. Output in json format where the key is the cluster, and the value is the list of input labels in that cluster. For each cluster, the value should only take labels from the user input. ONLY output the JSON object, and do not add any other text.
 \end{spverbatim}

\section{Experiment set-up}\label{app:experiments}
For our experiments, AI-assisted tools are used for coding and plotting.
\subsection{TopicGPT Example Topics}
\paragraph{Media Frame Corpurs} [1] Political: considerations related to politics and politicians, including lobbying, elections, and attempts to sway voters.

\paragraph{Astro Queries} [1] Knowledge seeking for specific facts: Questions about very specific pieces of information, such as characteristics, facts parameters of specific objects, phenomena, or processes.

\paragraph{ML Values} [1] Performance: A research value that show a specific, quantitative, improvement over past work, according to some metric on a new or established dataset.

\subsection{LLooM}
As \astro\ is the most difficult dataset to label, to not make LLooM disadvantageous, we use ``query type'' as the seed word for its label generation. 

\subsection{Incremental approach}
We set the number of text segments to run the generation module in the first iteration to be 32 and 48 for the rest of the iterations. The number of iterations that we run all three modules was set to 10 and after that, only merging and dropping modules will be run until all themes have more than one segment. When there are gold labels, another condition needs to be satisfied besides the 10 iterations condition, for the approach to stop running all three modules. The condition is that the approach needs to process at least 3 segments for each label before it starts to only run merging and dropping module. 
\subsubsection{Generation module}
We use the same template to \hicode\ prompt for fair comparison in the generation module.

\subsubsection{Merging module}
We use the following template of the system prompts for the merging module and pass the existing codebook as user prompts.

\begin{spverbatim}
"Synthesize the entire list of labels by clustering similar labels that are inductively labeled. 
The clustering is to finalize MEANINGFUL and INSIGHTFUL THEMES for {Goal of Inductive Coding}
You will be provided with an existing codebook. Now you need to cluster the codes into clusters and provide one higher level code for each cluster of codes.

Guidelines for Clustering:
- Analyze existing codes and their corresponding segments and cluster the existing codes into clusters with corresponding higher level codes representing the whole cluster.

The existing codebook will be provided as input following this example below:
1. <code1>

-> <segment labeled with code1>

2. <code2>

-> <segment labeled with code2>

...

n. <codeN>

-> <segment labeled with codeN>

Provide your answers following this output format example below:
Ans:
{{
"clusters": [
    {{
    "high_level_code": "Cyber Harassment",
    "original_codes": ["Online Harassment", "Cyberbullying"],
    "justification": "Both refer to aggressive online behavior; Cyberbullying is a subset but can be generalized."
    }}
]
}}

When no clustering is needed, answer N/A and provide your answer following this output format below:
Ans: N/A"
\end{spverbatim}

\subsection{Dropping module}
This module drops labels and their associated segments. The dropped segments will not get sampled for generation again. The intuition is to drop labels that are not merged or are not being generated again. In our experiments, we drop labels that are attached to only one segment if these labels do not attach to than one segment after 2 iterations.

\subsection{Reassignment}
The template for user prompting of the reassignment module is as follows.

\begin{spverbatim}
"{Goal of inductive coding} 
Analyze the following segment to identify the best label from the codebook that should be assigned to this segment. 

Segment will be given like below:
Segment:<segment text>

The existing codebook will be provided following this example below:
Codebook: 1. <code>, 2. <code>, ... , n. <code>

Provide your answers following this output format example below:
Ans: Cyber Harassment

Now given the existing codebook and the segment below, provide your answer following the format given above.
Codebook: <codebook>
Segment: <text_segment>"
\end{spverbatim}

\section{TopicGPT: Qualitative Analysis}
\label{sec:heatmap}

\begin{figure*}
        \centering
        \includegraphics[angle=90, width=0.3\linewidth]{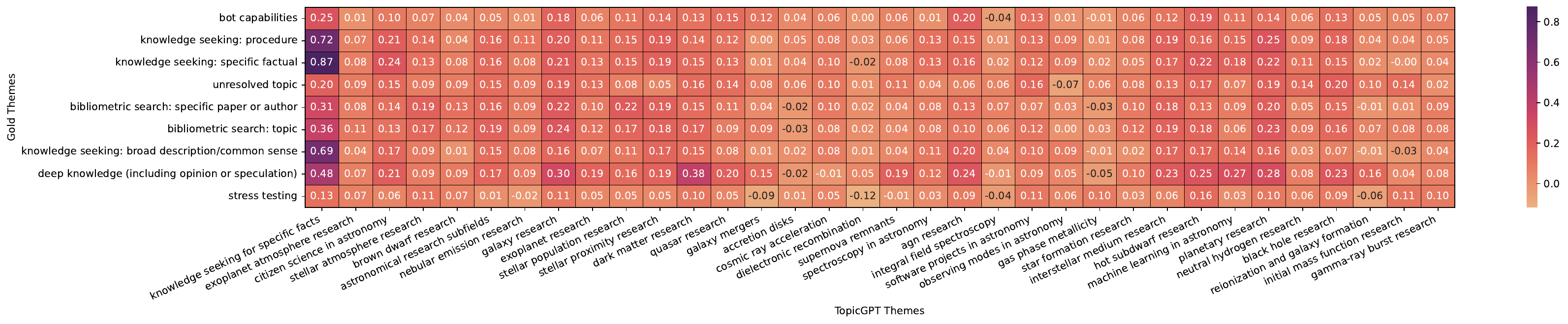}
        \caption{Heatmap comparing themes predicted by TopicGPT (horizontal) with human-labeled themes (vertical) for \astro. Each box contains the estimated similarity score between the predicted and gold theme, with darker colors signifying higher similarity.}
        \label{fig:automated-topicgpt}
\end{figure*}

In \cref{fig:automated-topicgpt} we provide a heatmap comparing the themes for \astro\ predicted by TopicGPT with the human-labeled themes, analogous to \cref{fig:heatmap}.

\section{Supplemental Materials for Model Ablation Study}
\label{sec:app_abalations}
Here we provide more details about the model ablation study in \cref{sec: ablation}. \cref{tab:generation-abalation} compares different models used for ``Label Generation''. \cref{tab:clustering-ablation-llama} and \cref{tab:clustering-ablation} compares different models used for ``Hierarchical clustering'' where the former used \llama{} in the generation stage and the latter used \mini{}.

\begin{table}[htb!]
\centering

\resizebox{\linewidth}{!}{
\begin{tabular}{lrr}
\toprule
  gen model & Precision & Recall\\
  \midrule
 \texttt{llama-3.2-3b} &  0.66 & 0.23\\
 \llama & 0.54 & 0.22\\
 \texttt{gpt-4o-mini}  & 0.68 & 0.26\\
 \texttt{claude}  & 0.63 & 0.30\\
 \texttt{mixtral-large} & 0.53 & 0.26\\
 \bottomrule

\end{tabular}}
\caption{
Clustering using \mini\ with generation results from different LLMs on \values\ data.}
\label{tab:generation-abalation}
\end{table}

\begin{table}[htb!]
\centering

\resizebox{\linewidth}{!}{
\begin{tabular}{lrr}
\toprule
  cluster model & Precision & Recall\\
  \midrule
 kmeans &  0.50 & 0.30\\
 \llama\ &  0.47 & 0.50\\
 \texttt{gpt-4o-mini} &   0.54 & 0.22\\
 \texttt{gpt-4o} & 0.55 & 0.33\\
 \texttt{claude} &  0.58 & 0.27\\
 \texttt{mixtral-large} &  0.58 & 0.21\\
 \bottomrule

\end{tabular}}
\caption{
Clutering using kmeans, \llama\, \mini, \texttt{gpt-4o}, \texttt{claude}, \texttt{mixtral-large} with generation results from \llama\ on \values{} data.
}
\label{tab:clustering-ablation-llama}
\end{table}

\begin{table}[htb!]
\centering

\resizebox{\linewidth}{!}{
\begin{tabular}{lrrr}
\toprule
  cluster model& \#clusters & Precion & Recall\\
  \midrule
 kmeans & 50 & 0.64 & 0.34\\ 
 \texttt{gpt-4o-mini} & 22 & 0.68 & 0.26\\
 \texttt{gpt-4o} & 59 & 0.55 & 0.40\\
 \texttt{claude} & 21 & 0.66 & 0.26\\
 \texttt{mixtral-large} & 53 & 0.54 & 0.41\\
 \bottomrule

\end{tabular}}
\caption{
Clustering using kmeans, \llama\, \mini, \texttt{gpt-4o}, \texttt{claude}, \texttt{mixtral-large} with generation results from \mini\ on\values{} data.
}
\label{tab:clustering-ablation}
\end{table}

\section{Supplemental Materials for Case Study}\label{app:case-study}
\begin{figure}
    \centering
    \includegraphics[width=\linewidth]{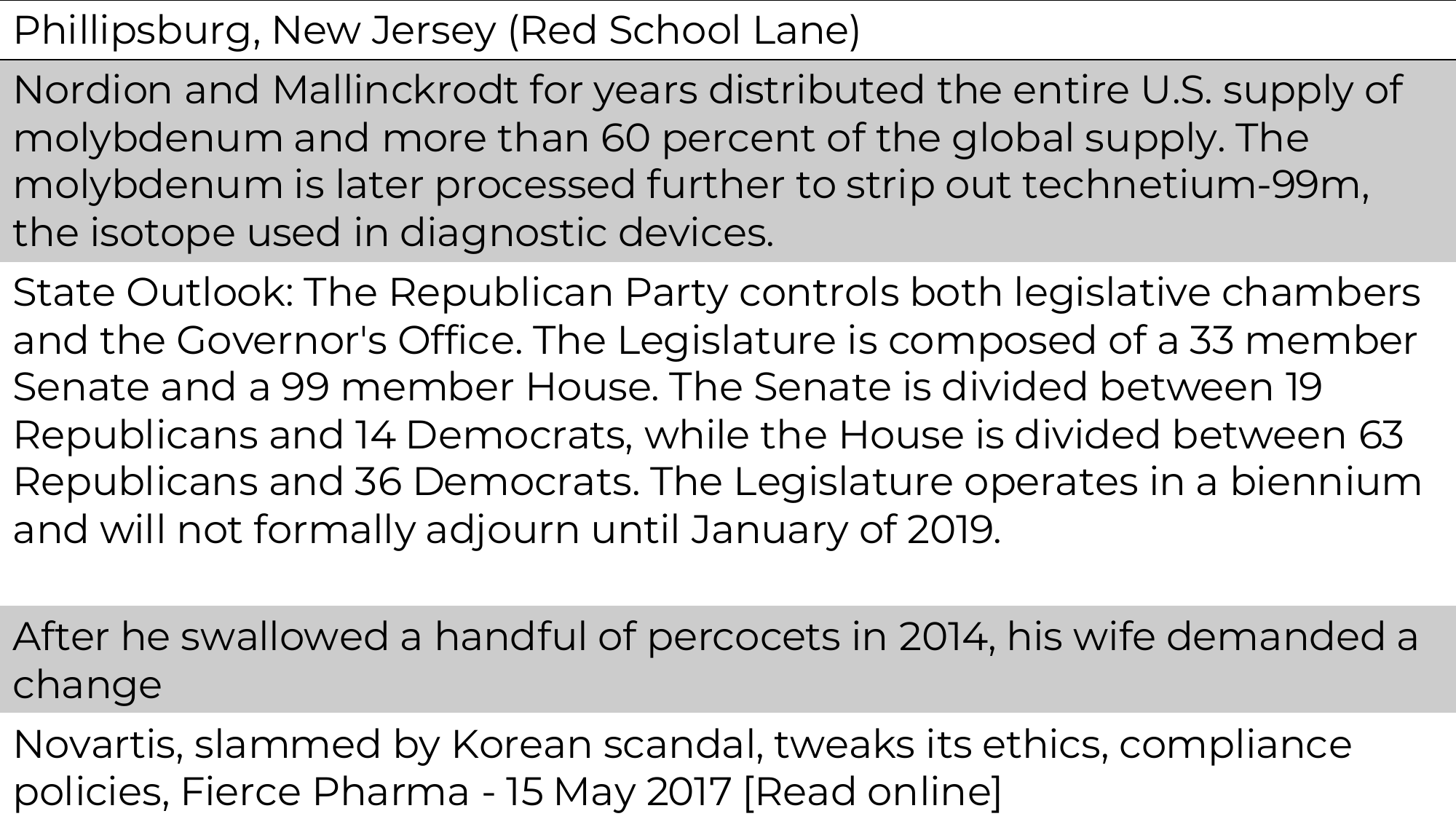}
    \caption{Examples of segments labeled as irrelevant segments.}
    \label{fig:irrelevant-quote}
\end{figure}

We provide some examples that are labeled as ``irrelevant'' in the label generation stage in \cref{fig:irrelevant-quote}.

\begin{figure*}
    \centering
    \includegraphics[width = \linewidth]{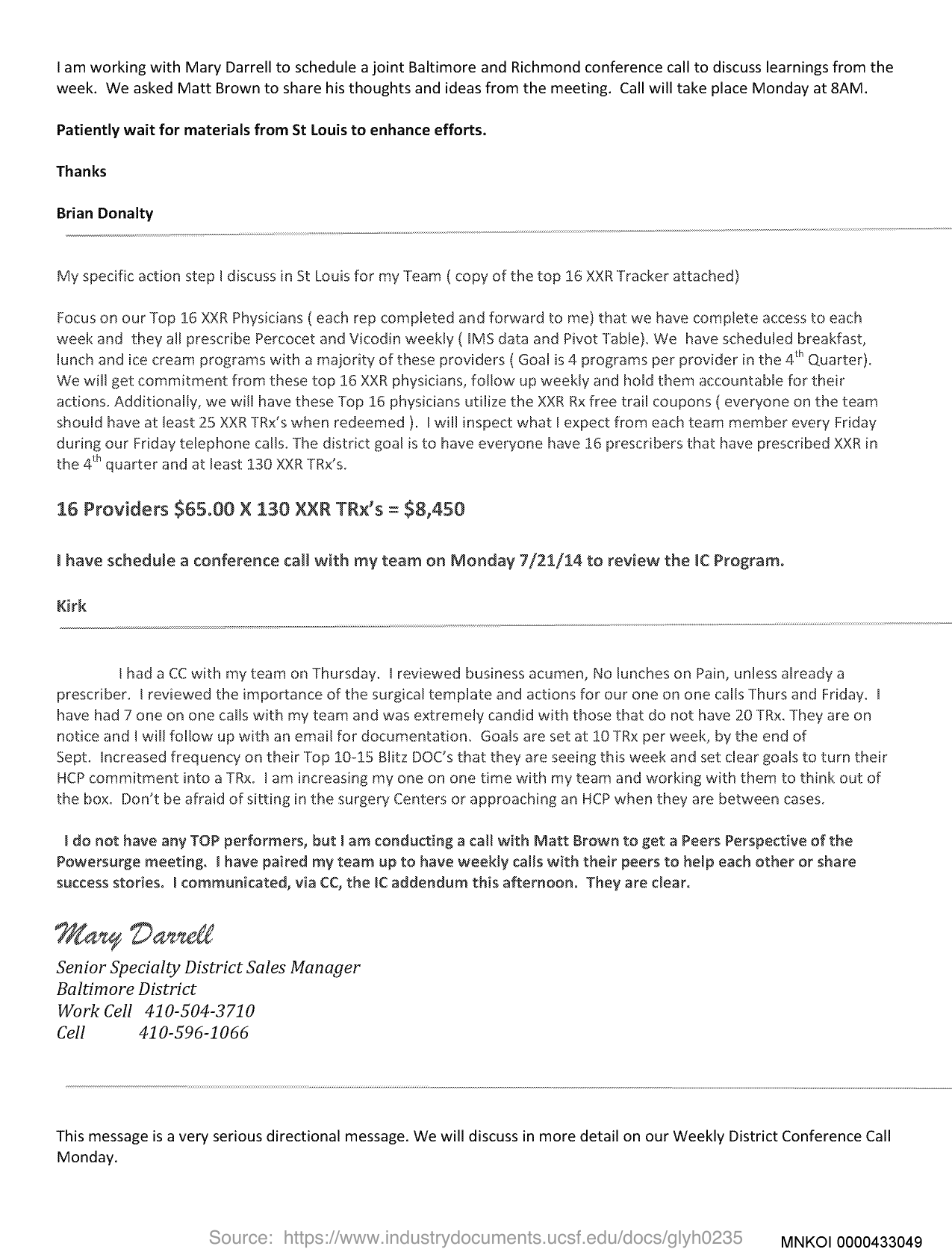}
    \caption{PDF example taken from OIDA file with id \texttt{glyh0235}}
    \label{fig:oida-pdf}
\end{figure*}

\begin{figure*}[htb!]
    \centering
    \includegraphics[width = \linewidth]{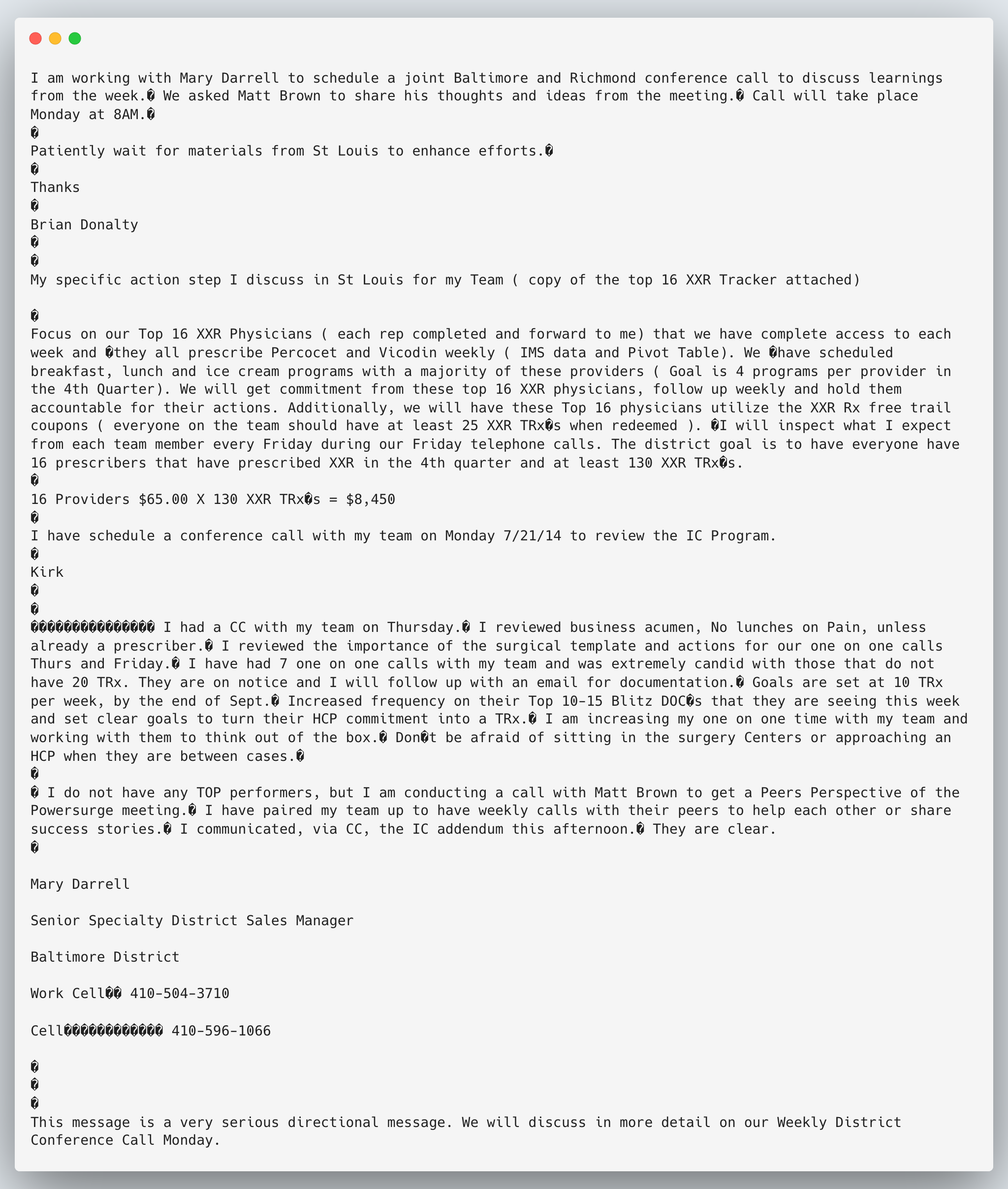}
    \caption{OCR example taken from OIDA file with id \texttt{glyh0235}}
    \label{fig:oida-ocr}
\end{figure*}

\begin{figure*}[htb!]
    \centering
    \includegraphics[width=\linewidth]{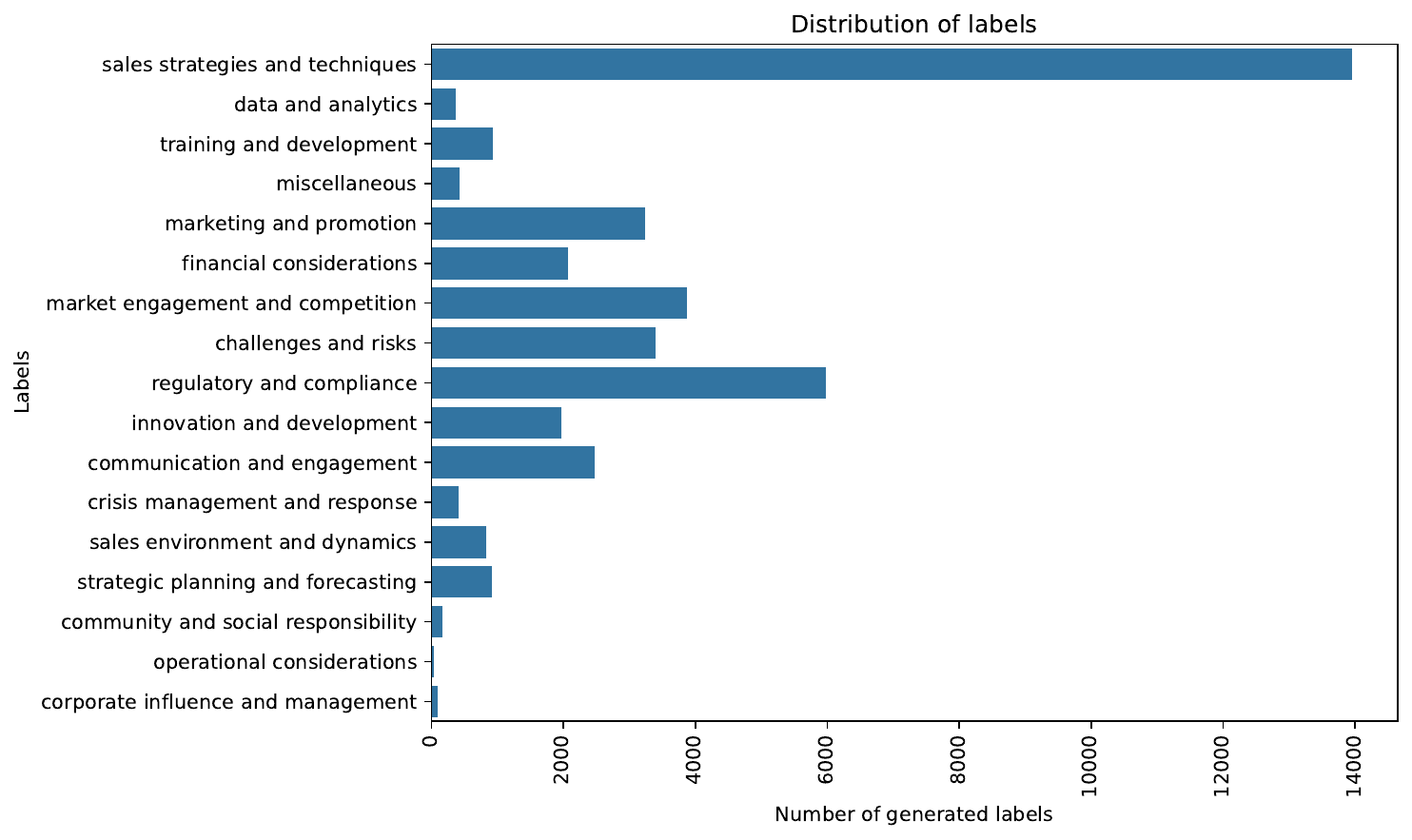}
    \caption{Distribution of final 17 themes on the OIDA sales contest data.}
    \label{fig:salescontest_label_distribution}
\end{figure*}

Moreover, \cref{fig:oida-pdf} and \cref{fig:oida-ocr} show the examples of the PDF and its OCR file~(with reduced new-line characters for better visual display).
\cref{fig:salescontest_label_distribution} shows the distribution of themes from the OIDA sales contest data. 

Finally, as mentioned in \cref{sec:case_study}, we provide the pairs of labeled text segments here for ``anticipating lost revenue'', ``identifying lost prescriber opportunity'', and ``anticipating opposition to expansion''.

\paragraph{Anticipating lost revenue}
\begin{spverbatim}
[Text Segment] 
Pfizer's investors are waiting the approval of new drugs to replace lost revenue and profit from the company's Lipitor (atorvastatin calcium), which will soon lose its patent protection. Pfizer saw \$10.7 billion in revenue from the drug last year, and analysts expect the company will have to replace about \$9.5 billion. Seeking Alpha published a look at the company's pipeline for candidates to potentially replace lost Lipitor revenue, focusing on new drugs that have the most financial potential. The drugs discussed are apixaban, an anticoagulation medication developed in partnership with Bristol-Myers Squibb; tofacitinib for treating immunological diseases including rheumatoid arthritis, inflammatory bowel disease and psoriasis; bapineuzumab for Alzheimer's disease; tanezumab for chronic pain; and crizotinib for non-small cell lung cancer.

[Label] 
- Patent protection loss strategy
- Pipeline development strategy
- New drug replacement strategy
- Revenue replacement strategy
- Investor expectation management
- Anticipating lost revenue
\end{spverbatim}

\paragraph{Identifying lost prescriber opportunity}

\begin{spverbatim}
[Text Segment]
a. Reason. Pete's top Exalgo prescriber stopped writing completely. This physician accounted for 35-40 prescription per month or around 120 a quarter. Pete has added several new prescribers but as you can see from the last couple of quarters he has not finished above 85-90\% quota attainment and his quota's have consistently increased the most in the district.
[Label]
- Identifying Lost Prescriber Opportunity
- Focusing on New Prescribers
- Quota Attainment Pressure

[Text Segment]
It's frustrating to see 75\% of the District's scripts coming from 3 territories. We had some territories with major losses this week which helps to explain our less than desirable results. Please take the time to review your data to find the physician's that have dropped off within your respective territories. We all know Exalgo offers significant advantages over other long acting opioids, we need to get our fair share of the scripts that are being written
[Label]
- Territory Performance Review
- Identifying Lost Physician Accounts
- Highlighting Product Advantage
- Script Share Focus

[Text Segment]
According to IMS we have had 46 different prescribers of XXR in the last 13 weeks. As we focus in on winning this district contest, getting each one of those prescribers to write again in Q4 will be critical. So far, only 13 of those prescribers have prescribed XXR in the first 2 reporting weeks of Q4. That means we have 33 others customers that have previously prescribed that we need to make sure prescribe again this quarter
[Label]
- Winning District Contest Focus
- Reactivating Prescribers Key Strategy
- Identifying Lost Prescribers

\end{spverbatim}

\paragraph{Anticipating opposition to expansion}

\begin{spverbatim}
[Text Segment]
Or if a Democratic president is elected and Republicans maintain control of Congress, Obamacare may remain the law of the land, but continue to limp along in heavily red states where antipathy runs so deep that state lawmakers continue to shun the expansion. The fact that states' costs will slowly rise over the next few years could further erode support. Electoral losses by pro-expansion lawmakers, which include a handful of Republicans, could have the same effect.

[Label]
- Anticipating opposition to expansion
- Potential electoral consequences 
- State lawmakers shun expansion
- Rising costs erode support
\end{spverbatim}

\end{document}